\documentclass[conference,10pt]{ieeetran} 

\usepackage{times}
\usepackage{mathtools}
\usepackage{url}

\usepackage{enumitem}
\usepackage[normalem]{ulem} 
\usepackage{algorithm}
\usepackage[noend]{algpseudocode} 
\usepackage{algorithmicx}

\usepackage{etoolbox}
\usepackage{tikz}
\usepackage{subfig}
\usetikzlibrary{tikzmark}
\usetikzlibrary{calc}

\newcommand{\ALGtikzmarkcolor}{black}
\newcommand{\ALGtikzmarkextraindent}{4pt}
\newcommand{\ALGtikzmarkverticaloffsetstart}{-.5ex}
\newcommand{\ALGtikzmarkverticaloffsetend}{-.5ex}
\makeatletter
\newcounter{ALG@tikzmark@tempcnta}

\newcommand\ALG@tikzmark@start{%
	\global\let\ALG@tikzmark@last\ALG@tikzmark@starttext%
	\expandafter\edef\csname ALG@tikzmark@\theALG@nested\endcsname{\theALG@tikzmark@tempcnta}%
	\tikzmark{ALG@tikzmark@start@\csname ALG@tikzmark@\theALG@nested\endcsname}%
	\addtocounter{ALG@tikzmark@tempcnta}{1}%
}

\def\ALG@tikzmark@starttext{start}
\newcommand\ALG@tikzmark@end{%
	\ifx\ALG@tikzmark@last\ALG@tikzmark@starttext
	\else
	\tikzmark{ALG@tikzmark@end@\csname ALG@tikzmark@\theALG@nested\endcsname}%
	\tikz[overlay,remember picture] \draw[\ALGtikzmarkcolor] let \p{S}=($(pic cs:ALG@tikzmark@start@\csname ALG@tikzmark@\theALG@nested\endcsname)+(\ALGtikzmarkextraindent,\ALGtikzmarkverticaloffsetstart)$), \p{E}=($(pic cs:ALG@tikzmark@end@\csname ALG@tikzmark@\theALG@nested\endcsname)+(\ALGtikzmarkextraindent,\ALGtikzmarkverticaloffsetend)$) in (\x{S},\y{S})--(\x{S},\y{E});%
	\fi
	\gdef\ALG@tikzmark@last{end}%
}

\apptocmd{\ALG@beginblock}{\ALG@tikzmark@start}{}{\errmessage{failed to patch}}
\pretocmd{\ALG@endblock}{\ALG@tikzmark@end}{}{\errmessage{failed to patch}}

\usepackage{subfig}

\usepackage[bookmarksnumbered,unicode,colorlinks=true,linkcolor=red,urlcolor=blue,citecolor=blue]{hyperref}
\usepackage{cleveref}

\crefformat{section}{\S#2#1#3} 
\crefformat{subsection}{\S#2#1#3}
\crefformat{subsubsection}{\S#2#1#3}

\usepackage{listings}
\usepackage{xcolor}

\definecolor{codegreen}{rgb}{0,0.6,0}
\definecolor{codegray}{rgb}{0.5,0.5,0.5}
\definecolor{codepurple}{rgb}{0.58,0,0.82}
\definecolor{backcolour}{rgb}{0.95,0.95,0.92}

\lstdefinestyle{mystyle}{
  commentstyle=\color{codegreen},
  keywordstyle=\color{magenta},
  numberstyle=\tiny\color{codegray},
  stringstyle=\color{codepurple},
  basicstyle=\ttfamily\footnotesize,
  breakatwhitespace=false,         
  breaklines=true,                 
  captionpos=b,                    
  keepspaces=true,                 
  numbers=left,                    
  numbersep=5pt,                  
  showspaces=false,                
  showstringspaces=false,
  showtabs=false,                  
  tabsize=2
}
\usepackage{balance}
\usepackage{hyperref}
\usepackage{authblk}
 \usepackage{cite}

\def \myname{\textsc{GnnBench}}

\begin{document}
%
\title{\Large \bf {\myname}: Fair and Productive Benchmarking for Single-GPU GNN System}

\author{Yidong Gong}
\author{Pradeep Kumar}
\affil{William \& Mary}



%


\maketitle
\pagestyle{plain}

%
\begin{abstract}

We hypothesize that the absence of a standardized benchmark has allowed several fundamental pitfalls in GNN System design and evaluation that the community has overlooked.
In this work, we propose {\myname}, a plug-and-play benchmarking platform focused on system innovation. {\myname} presents a new protocol to exchange their captive tensor data, supports custom classes in System APIs, and allows automatic integration of the same system module to many deep learning frameworks, such as PyTorch and TensorFlow. 
To demonstrate the importance of such a benchmark framework, we integrated several GNN systems. Our results show that integration with {\myname} helped us identify several measurement issues that deserve attention from the community.

\end{abstract}
\section{Introduction} \label{sec.intro}

Many applications have become data-driven today where 
benchmarking is critical in advancing systems innovations.  We have seen benchmarking-driven research recently; popular examples include Graph500~\cite{graph500}, LDBC ~\cite{LDBC}, LinPack~\cite{LinPack}, SPEC~\cite{SPEC}, and Large Scale Visual Recognition Challenge 
which have advanced research in graph analytics, HPC, computer vision tasks, etc. Such benchmarks have been written based on the insights of the corresponding application domain. E.g., Graph500, and LDBC have advanced the study of system optimizations of the irregularity prevalent in graph analytics. 

Graph neural network (GNN) models, e.g., GCN~\cite{gcn17iclr}, GAT~\cite{gat18iclr}, GIN~\cite{xu2019powerful}, and others~\cite{gatedgraph2017,zhang18, graphsage17} have become an important data-driven application.  They all use \textit{sparse matrices} or \textit{graphs} as the data model to solve problems such as in social networks, chemistry, biology, recommendation system, scene description generation, knowledge bases, etc~\cite{fan2019graph, zitnik2018modeling,hamilton2018embedding,ying2018graph,krahmer2003graph,perera2017recent,rgcn2018,wang2020entity}. 
However, prior benchmarking efforts, such as Graph500 or LDBC cannot be used for benchmarking system innovations in GNNs because GNNs a) involve forward and backward computations unlike forward-only classical graph analytics, b) need to store the output generated in the forward pass for backward computation, c) use vector features for vertices and edges, and {d) relies on DL frameworks, such as PyTorch, TensorFlow to perform automatic differentiation.

\noindent \textbf{Motivation.}
The goal of GNN system research is to improve the runtime and memory consumption while keeping \textit{model semantics} the same, i.e., not changing the math behind training.
Yet, many system design and evaluation pitfalls in GNN have been observed~\cite{graphpy2024}. E.g., GNNAdvisor~\cite{wang2021gnnadvisor}, TC-GNN~\cite{wang2023tc}, etc. do not include the bias in their GCN model but keep it in their baseline system DGL~\cite{dgl2019} when comparing runtime. GNNAdvisor and Seastar~\cite{wu2021seastar} did not correctly call the back-propagation kernel for \textit{normalization by degree} computation, while TC-GNN removed it. A few GNN systems do not implement transposed SpMM (SpMM$^T$)~\cite{wang2023tc,chen2020fusegnn,huang2021understanding}, and either rely on a simulated forward computation, or substitute SpMM in place of SpMM$^T$. Many others have also removed other layers such as dropout, etc. 
Finally, many current single-GPU GNN systems exclusively rely on smaller datasets where training time is dominated by framework overhead~\cite{graphpy2024} thereby their evaluation is not reliable. However, when we try to run some of them on mid-size datasets, they crash due to runtime issues, missing backward computation, etc., limiting future works to benchmark against them.


Systems suffering from pitfalls must be corrected to use them as baselines for future research comparison. However, we argue that fixing the pitfalls manually is not the correct approach, but rather we argue for a more systematic approach of providing a comprehensive benchmarking platform. Indeed, as the goal is to optimize the semantic-preserved model, a more systematic approach is possible and can also be \textit{productive} for future works which can minimize efforts to avoid those pitfalls. Specifically, GNN systems can avoid the additional effort to build front-end, integration to DL Framework, etc. as the aim is never to innovate this part. 
Further, for \textit{fair evaluations}, one has to ensure that the developed system has the same model specification as the other baselines. A benchmarking system ensures such fairness automatically. \textit{However, there is no existing work that can achieve this.}

\noindent \textbf{Vision and Challenges.}
Our vision is to allow system researchers to productively plug their kernel (system) binary into the proposed benchmarking suite so that researchers can focus on system innovations without worrying about integration, evaluation pitfalls, or fair comparison. 
Thus a standardized GNN benchmark should provide all the front-end code (model specification) and specify well-defined and stable interfaces (henceforth, the \textit{System API}) between front-end (Python) and back-end (system) modules, so that system binary can become a plug-and-play module.



This paper focuses on challenges in designing such APIs for GNN benchmarking when many system innovations do not allow it to be stable including when a researcher proposes a custom storage format. 
For example, the PyTorch plugin development environment only provides tensors and scalar data types to be specified in the System APIs, but not a custom class, such as a `graph'. This forces researchers to model graphs using tensors.
E.g., a CSR format can be modeled using two one-dimensional (1-D) tensors representing row offsets and column indices. However, the custom format proposed by GNNAdvisor~\cite{wang2021gnnadvisor}, Seastar~\cite{wu2021seastar}, TC-GNN~\cite{wang2023tc}, and others~\cite{huang2021understanding,dgNN} etc. cannot fit in two tensors and may require 3 or more 1-D tensors. Hence, any new custom format requires continuous changes to the signature of System APIs.

\begin{figure}[t]
\includegraphics[scale= 0.64]{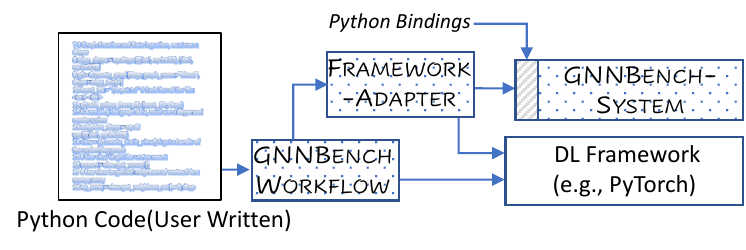}
\vspace{-6pt}
\caption{\small {\myname-System} is DL framework agnostic, while integration is done at Python level using {Framework-Adapter}. An independent {\myname-System} has no interface-level limitation.} 
  \label{fig-arch}
\end{figure}

 \noindent
\textbf{Contribution}. 
We propose {\myname}, a standardized GNN benchmark, built on a modular and extensible platform to solve the above-discussed challenges.
Fig.~\ref{fig-arch} shows a high-level overview of {\myname}, which separates the functionality into two modules. {\myname-System} is the plug-and-play system module and  a common {\myname-Workflow} provides the front-end code for the GNN model interacts with {\myname-System} through system APIs. 

\noindent $\bullet$\textbf{Stable System APIs.} 
We propose to convert a DL Framework's private and captive tensor data structure to a \textit{zero-copy} C-like flat array. The latter is passed to {\myname-System} to make it \textit{completely independent} of DL Framework's plugin development environment while allowing its full integration at the Python level by {Framework-Adapter}.
This independence, in turn, makes the system APIs as powerful as between C/C++ and Python, thereby allowing custom classes, such as  `graph', in the system API signatures leading to stability even when storage format changes. The developed protocol is generic and can be used by many other systems (\cref{sec.exchange}). 



\noindent $\bullet$
\textbf{Productivity and Fairness.}
Common GNN model front-end (part of {\myname-Workflow}), and the independent nature of {\myname-System} through stable System APIs let researchers prototype the system aspect of GNN quickly, and get immediate feedback by evaluating it through {\myname}.
Further, we propose a simple domain-specific language (DSL) that uses the kernel signature to generate all the critical system integration code, Python binding, autograd code, and the skeleton of the kernel implementation.


\noindent $\bullet$
\textbf{DL Framework Agnostic.}
The complete independence of {\myname-System} allows it to be integrated into many DL frameworks, such as PyTorch and TensorFlow. 
{\myname-system} can be plugged into any of the frameworks.

\noindent $\bullet$
\textbf{New Artifacts.}
Many original GNN systems not only have pitfalls but also cannot run due to code integration issues, unknown memory corruption, or missing backward computation.
{\myname} can productively integrate with these GNN systems (GNNAdvisor~\cite{wang2021gnnadvisor}, TC-GNN~\cite{wang2023tc}, GE-SpMM~\cite{gespmmsc2020}, Huang et al~\cite{huang2021understanding} and Cusparse) or new ideas, while automatically fixing their pitfalls, if any.
These new artifacts serve not only as a useful baseline for future research but also as state-of-the-art as we show during evaluations. 

\noindent 
$\bullet$ \textbf{New Insight from Evaluation.} A benchmark system's strength lies in the insight derived from evaluation.
To this end, our evaluation shows that many pitfalls are automatically fixed by their integration. Further, all integrated systems take the same runtime and achieve the same speedup over DGL when the dataset is small showing the current reliance on smaller datasets is not a good practice. For mid-level datasets, 
GE-SpMM, one of the earliest GNN systems with no workload balancing, outperforms many other systems with workload-balancing techniques.

The memory consumption evaluation shows that DGL not only consumes memory due to some inefficient design but also suffers from huge framework-memory overhead-- memory that is not consumed by any GNN kernels or graph storage. 

The runtime evaluation shows that Cusparse integrated {\myname} is still the fastest for the complex GAT model even on top of a highly fused state-of-the-art system, dgNN. The latter is slower by 45.77\% while saving a modest memory of around 16.44\% for GAT over our Cusparse integrated {\myname} at the cost of huge efforts in implementing custom fused kernels. The results force us to revisit the advantage of the kernel fusion model.

\noindent \textit{Organization}: We discuss background in \cref{sec.background}, challenges and related works in \cref{sec.mot}, detailed design in \cref{sec.overview}, evaluation in \cref{sec.exp}, and conclusion in \cref{sec.conclusion}.

\section{Background on GNN Computation} \label{sec.background}


\begin{figure}[t]
	\centering
	\vspace{-4mm}
	\includegraphics[scale=0.59]{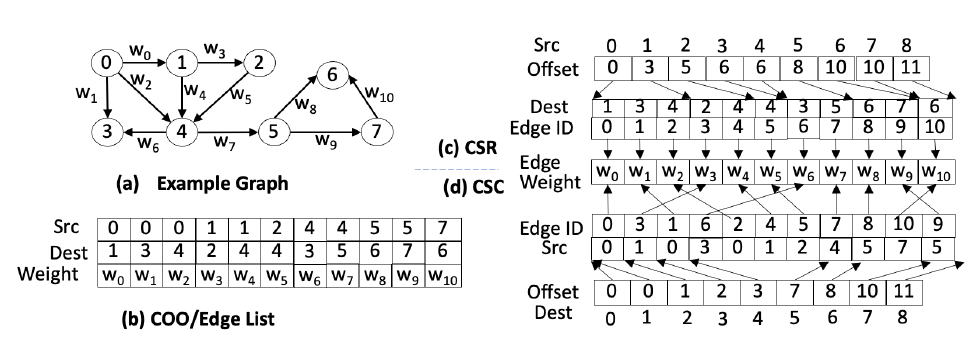}
	\caption{\small Graph representation in DGL: it has introduced an edge ID to the graph.} 
	\label{fig-format}
\end{figure}
%

A graph consists of vertex set V and the edge set E. W refers to the optional edge features. In GNN, vertices also keep features.
Fig.~\ref{fig-format} shows an example of such a graph and its corresponding representations. 
The \textit{compressed sparse row} (CSR) format stores all the neighbors ($N(v)$) of the vertex sequentially and uses the offset array to link the vertex and the first edge of its corresponding N(v). The reverse graph in the case of a directed graph also needed to be stored; in this case, CSR stores out-neighbors while \textit{compressed sparse column} (CSC) format stores the in-neighbors of each vertex. The \textit{degree} of a vertex is the length of N(v). We also use linear algebra terminology interchangeably to represent the mathematical operation behind GNN.

A GNN model is made up of model layers. 
 Each layer internally invokes the \textbf{operators}, which is implemented using the  
\textbf{Autograd} class: it contains forward() and backward() as its member function, which one can write to implement a custom operator. These two methods internally call one or more \textbf{kernels} for the actual computation which is implemented in high-performance languages, like CUDA.

To this end, \textit{SpMM} (sparse matrix matrix multiplication) ($Y \leftarrow A_{W}X$) is a widely known \textit{sparse kernel} used in GNNs. Its backward computation is \textit{ SpMM$^T$} ($\delta X \leftarrow A^T \delta Y$) and Sampled dense dense matrix multiplication (\textit{SDDMM}) ($\delta W \leftarrow A \odot (X^T \delta Y)$). 
Here $A_W$ represents the graph (or sparse matrix), where each edge has features, and is collectively represented as $W$ (\textit{edge-level tensor}). $A$ is just the graph topology, i.e., its edge features are treated as 1.0. $\delta$ represents gradients of the tensors. We denote transposed SpMM as  $SpMM^T$. $X$ and $Y$ are \textit{vertex-level tensors}.




Deep learning (DL) frameworks, such as TensorFlow \cite{abadi2016tensorflow}, PyTorch \cite{paszke2019pytorch} 
, etc. create a \textbf{computation graph} or \textit{directed acyclic graph} (DAG) of operators from the model specification.
In the \textbf{forward} pass of the model training, the computation graph calls a sequence of \textit{kernels} from the input layer to the output layer to derive the prediction value, which is compared with the ground truth label to calculate the loss and plug the loss into the backward procedure. 
Then, in the \textbf{backward} computation, the loss is used to calculate the gradient by calling a sequence of kernels from the output layer to the input layer. It uses the chain rule of gradient back-propagation to update the model parameter.

\textit{Normalization by degree array} also has many variants. The simplest variant is where we divide vertex-level features directly by degree, in which vertices with zero degrees need to be set as 1.0 before division. Other variants perform division by degree before and after SpMM using the square root of source and destination degree arrays. Hence, many systems keep a separate array to store the degree array.

\noindent \textbf{GNN System and Stable API.}
The system code containing the kernel implementation is referred to as the \textit{GNN system} and its kernels as \textit{system API} after it is exported as a Python package. 
The system API is called \textit{stable} if its signature does not require changes when its input parameter, such as a graph storage format changes its format. The model code is usually written in Python.

\section{Goals, Challenges, and Related Works} \label{sec.mot}
A fair and productive benchmarking serves past and future systems both. Specifically, several GNN systems suffer from system design and evaluation pitfalls~\cite{graphpy2024}, hence we must correct their issues to use them as baselines for future research comparison. The proposed {\myname} aims to have a more systematic approach to benefit future works and help perform fair evaluations so that systems innovation can be advanced. 
{\myname} relied on a modular design to enable researchers to plug-and-play their {\myname-System} and propose their innovation as long as it supports the stable system APIs.  
We now set its key objectives in this regard.

\noindent
\textbf{Provide Sufficient Usability and Extensibility.}
{\myname} should have \textit{stable system APIs} to integrate diverse system ideas and allow productive benchmarking. Moreover, it should provide researchers with sufficient space for system innovation without imposing limitations. 
Furthermore, various DL Frameworks, such as TensorFlow and PyTorch, remain popular, and {\myname} should not force researchers to rely on a specific DL framework and should either offer integration codes or generate them automatically to speed up the prototyping process based on any custom need of a system.


\noindent
\textbf{Assistance in Identifying and Correcting Accuracy Issues.}
Identifying accuracy issues in many original GNN systems that suffer from pitfalls~\cite{graphpy2024} is challenging due to the complexity of the systems, e.g., errors may arise due to various issues in the requirement understanding, model specification, kernel issues, or a mix of them. 
By providing a well-tested workflow implementation including saving state tensors, and passing several flags in the system API to indicate the right auxiliary operations, such as transpose, we remove the need to revisit the requirement understanding and model specification. Thus a correctly implemented kernel will show good accuracy if integrated to {\myname}.


\noindent
\textbf{Fair Benchmarking amid Low Framework Overhead}
A benchmarking framework should have very low overhead in terms of resources consumed, i.e., low \textit{framework-runtime overhead} while executing the framework code, and \textit{no memory overhead}.
In recent years, numerous GNN systems have been proposed, each employing different methods to integrate and implement their kernels within GNN model layers. This leads to differences in \textit{framework overhead}, which interferes with the runtime measurements, specifically affecting measurements on smaller datasets~\cite{graphpy2024}. 
Further, popular baselines such as DGL introduce \textit{framework-memory overhead} (see \cref{sec.exp}), where a major portion of the memory is consumed by the DGL framework and not by graph storage or kernel computation. 
To this end, {\myname} must minimize framework overhead, and introduce no framework-related memory overhead so that systems integrated to {\myname} can perform fair performance comparisons, while other correct systems developed outside of {\myname} cannot have an unfair advantage.

\subsection{Challenges and Related Works} \label{sec.challenge}
We discuss that prior GNN systems have different sets of system APIs despite their semantics being the same, thereby presenting challenges in achieving stable APIs. We analyze their root cause and discuss that other systems cannot serve as benchmarking systems.

\subsubsection{Instability due to Graph Format}
Pytorch's Plugin interfaces accept only tensor type, a specific smart pointer (c10::intrusive$\_$ptr), and built-in scalar types. Hence,
\textit{it cannot accept a custom graph object}, hence different systems require different numbers of 1-D arrays to model the graph.
GNN systems relying on COO formats~\cite{pytorchg2019,spektral,battaglia2018relational,grattarola2021graph} have modeled graph storage using either two 1-D tensors of source and destination vertex ID or one tensor where each element keeps source and destination vertex IDs together.
For CSR format, TLPGNN~\cite{fu2022tlpgnn} use two 1-D tensors to represent the offset array and neighbor array (Fig.~\ref{fig-format}) of CSR.
GNN systems based on custom format introduce additional metadata to the format.
E.g., GNNAdvisor's neighbor grouping adds a 1-D array of size greater than $|V|$ to the CSR format. 
Similarly, Sputnik~\cite{gale2020sparse} and Seastar~\cite{wu2021seastar} sort the vertices based on their degree, which needs an additional 1-D array to keep explicit vertex IDs. 
TC-GNN~\cite{wang2023tc} uses a compressed tiled structure that requires five 1-D arrays to store the new storage format. Similarly, tile-based storage, such as  Apst~\cite{hong2019adaptive} requires different counts of 1-D arrays to model the storage format. 

We did not find any prior works that use the smart pointer approach. However, we found it complex to implement as implementation requires
a better understanding of the interface limitation and special conventions that are not very explicit in the documentation. It took us 6 months due to complex and unclear documentation (see appendix for timeline).
TensorFlow's plugin also restricts passing custom objects, unless additional steps are performed using its \textit{experimental.extensionType} plugin, which is conceptually similar to the smart pointer approach of the PyTorch plugin.

\subsubsection{Instability due to Kernel Variants}
GNN operations force the presence of additional tensors or graph storage metadata in its signature depending on the strategy followed by different GNN systems.
For SpMM, weight implies features (scalar or vector) on edges, hence an additional edge-level tensor may or may not be needed in the system API based on SpMM type.
Similarly, SpMM$^T$ have forced many systems to keep additional metadata in the graph storage, e.g., DGL, Seastar, and FeatGraph have introduced explicit edge ID arrays in their CSR and CSC storage, which is then passed to SpMM and SpMM$^T$. However, Cusparse natively provides SpMM and SpMM$^T$ without requiring edge ID indirection. Hence, if one is willing to build GAT using the Cusparse library, its system API for these two kernels will differ.
dgNN~\cite{dgNN} has introduced additional metadata on top of their CSR format, called permute (1-D tensor) to solve problems that arise due to kernel fusion, and requires passage of this information in their backward System API, but not in the forward API.

For SDDMM, different researchers have used different storage formats compared to their SpMM implementation. DGL relies on the COO format, Seastar uses the CSR/CSC format with explicit edge ID, etc. dgNN on the other hand, uses plain CSR format for this kernel.

\subsubsection{DGL's Framework overhead and Complex Integration}
DGL suffers from framework overhead for runtime and memory consumption (\cref{sec.exp}). Hence systems integrated into it also have the same issues. The system integration process is unclear and undocumented. Hence prior works have followed different approaches after learning its internal.
GE-SpMM~\cite{gespmmsc2020} provides the source code files to replace the same-named DGL source files. However, those named files are now missing in DGL due to its complete code re-organization, a trait of a monolithic system. 
So, questions remain on the modality of integration. 

Supporting a custom graph storage needs to change almost the whole DGL backend, 
hence spending additional effort may be useless. History indicates that such types of optimizations~\cite{huang2021understanding,wang2021gnnadvisor,fu2022tlpgnn,spektral} have rather relied on plugin environment, developed their own model code, system APIs, and system code despite the interface-level limitation.
One of the authors of a prior GNN system aptly described the limitations of working with DGL: \textit{``We didn't use DGL, because some of its design is inefficient but is also embedded in the system design in a complex way''} 
confirming our observation on DGL.

\subsubsection{Other Platforms}
DNN-related benchmark platforms \cite{mlperf, TBD, dnnmark, Fathom, DAWNBenchA} have been proposed recently, but the presence of the graph makes them unusable for GNNs.
GNN benchmarking has remained an understudied problem. We are not aware of any systems that have focused on this problem. GNNMark~\cite{baruah2021gnnmark} 
is not designed as a standardized framework to integrate system kernels, making it impossible to integrate with any system-level GNN optimizations.

Pytorch-Geometric(PyG)~\cite{pytorchg2019}, relying on the PyTorch plugin, suffers from the limitations that a plugin system suffers.

\section{{\myname} Approach} \label{sec.overview}
\noindent
\textbf{Overview.} {\myname} innovates on interface design to offer stable system APIs. 
We make {\myname-System} fully \textit{independent} of the DL framework into which it is integrated so that system APIs avoid restrictions imposed by the DL Framework's plugin method. Thus we can support what is offered natively between Python and C/C++, including the custom 'graph' class. 
The challenge is how such an independent {\myname-System} integrates with a DL Framework, and receives its private and captive tensor data.
To this end, our proposed \textit{producer-only DLPack protocol} (\cref{sec.protocol}) is a simpler and elegant approach to sharing DL frameworks' tensor to third-party libraries (e.g., {\myname-System}). 
Taking advantage of the generality of the proposed protocol and independence of {\myname-System}, \textit{Framework-Adapter} integrates the same {\myname-System} with PyTorch and TensorFlow.


Table~\ref{table-kernel} shows fine-grained system APIs, and are discussed in \cref{sec.api} that they are comprehensive enough to cover all different kinds of GNN models to resolve the instability due to kernel variants. 
For other custom system APIs and to support unrestricted innovations, system APIs and integration codes are automatically generated by specifying the API signature using a simple DSL (\cref{sec.dsl}). These APIs require changes only when kernel fusion is attempted on system APIs thereby providing extensibility and productivity.

{\myname} uses other GNN systems' system APIs (SpMM and SDDMM) to build many GNN models, including GCN, GIN, and GAT. Other operators such as a linear layer, bias, normalization by degree array, etc. are common. 
It builds infrastructure to evaluate their performance and accuracy and provides a unified way to unit test all the kernels for debugging and individually benchmarking their performance. 
One clarification to note is that we did not modify the fundamental design of the kernels of prior works, minor accuracy-related fixes were made for pitfall-manifested systems. 
Their kernel implementation can still be viewed as their design. (\cref{sec.significance})

\begin{table}[t]
\small
\caption{\small {\myname-GNN} System APIs. Goal: a) Passing graph object as is, and remaining independent of DL framework; b) Their implementation must invoke the CUDA or device kernels internally.} 
\vspace{-6pt}
\centering 
\setlength{\tabcolsep}{1mm}
\begin{tabular}{|l|} 
    \hline 
    \hline
    status \hspace{2mm} SpMMv(g, \textit{in-tensor}, \textit{out-tensor}, eFn, Flag)\\
    status \hspace{2mm} SpMMve (g, \textit{in-tensor}, \textit{in-tensor}, \textit{out-tensor}, eFn, Flag)\\
    status \hspace{2mm} SpMMe(g, \textit{in-tensor}, \textit{out-tensor}, eFn, Flag)\\
    status \hspace{2mm} SDDMMvv(g, \textit{in-tensor}, \textit{out-tensor}, eFn, Flag)\\
    status \hspace{2mm} SDDMMve(g, \textit{in-tensor}, \textit{in-tensor}, \textit{out-tensor}, eFn, Flag)\\
    \hline
\end{tabular}
\label{table-kernel} 
\end{table}

\subsection{Zero-Copy Tensor Exchange} \label{sec.exchange}
To make {\myname-System} independent of the DL framework passing the private tensor object of the DL framework to {\myname-System} is the main task. To this end, we first discuss the current DLPack protocol~\cite{dlpack}. It should be noted that the protocol is not very well documented due to missing white paper, though some
discussion exists and links to frameworks that provide
implementations. Also, there is no formal name for the protocol, hence we take liberty to name it \textit{producer-consumer protocol} based on our understanding. We then present our proposed \textit{producer-only protocol}.

\subsubsection{\textbf{Existing Producer-Consumer DLPack Protocol}} \label{sec.oldproto}
DLPack makes the DL framework's captive tensor object interoperable with other frameworks.
It is an intermediate data structure, a \textit{container} that wraps the captive tensor's underlying memory pointer, size, shape, stride, etc. The protocol outlines two sets of APIs: \textit{producer APIs} and \textit{Consumer APIs}. 
Listing~\ref{list-spmm-orig} shows the usage of these APIs (not the implementation).
Producer APIs convert a producer-captive tensor (e.g. Pytorch tensor) to a Python capsule containing the pointer to the DLPack container object (Lines 7--8). The consumer, such as {\myname-System}, receives this capsule and converts its enclosed DLPack container to their private tensor object (Lines 10--11) before calling the System API. In this whole process, a zero-copy of the underlying memory is maintained.

\begin{lstlisting}[language=Python, caption= {\small Attempt \#1: Showing DLPack Producer-Consumer Protocol Usage: Tensors are allocated in Python, converted to DLPack tensors (capsule), and consumed to convert to {\myname-System's} captive tensor before passing them to the system API. This approach is complex, increases efforts, and is error-prone.}, label={list-spmm-orig}]
import GB_system as gbs
import torch as th
def gb_SpMMv(g, inTorchTensor, reduceFn, Forward):
  #Allocate output tensor here and not in Kernel.
  outTorchTensor = th.zeros(g.get_vcount(), featLen)
  # Use Producer API
  inDlpackTensor = th.to_dlpack(inTorchTensor)
  outDlpackTensor= th.to_dlpack(outTorchTensor)
  # Use Consumer API
  inGbsTensor = gbs.from_dlpack(inDlpackTensor)
  outGbsTensor= gbs.from_dlpack(outDlpackTensor)
  #Call  SpMMv System API.
  gbs.SpMMv(g, inGbsTensor, outGbsTensor, reduceFn, Forward)  
  return out_tensor 
\end{lstlisting}


It should be noted that the DLPack protocol is about managing the lifetime of the DLPack container, as the underlying memory space is still owned by the producer which, of course, cannot be deleted till the container is alive.   
The protocol lays out rules and mechanisms for the ownership transfer of DLPack containers. Initially, the producer keeps the ownership of the container after the producer API is invoked. After the consumer calls the consumer API, the ownership must be transferred to it. Once the consumer is done with tensor memory usage (e.g., kernel computation is done), it frees up the container object to inform the producer that the consumer will not access the tensor memory.

\noindent \textbf{Issues.}
We present two issues with this protocol. \textit{Firstly}, it is not clear from the documentation if a consumer system has to implement both (producer and consumer) APIs as systems have implemented both sets of APIs.
\textit{Secondly}, consumer-side of the protocol is complex,
where an independent developer faces the risk of not handling corner cases ~\cite{complex_comsumer}
contains some commentary. For example, the ownership of the resultant DLPack container object is determined by a string field within it with two values: ``dltensor'' and ``used\_dltensor''.
The former value implies that the producer is still the owner, but the latter value implies that the consumer is the owner, and hence the consumer API must change ``dltensor'' to ``used\_dltensor''. 
This mechanism itself is a candidate for data-race as set and test for string values are not atomic; who should allocate and free these two string values introduces the same memory management problem: the solution is to rely on statically allocated memory so that they do not require freeing.

Further, freeing the consumer-owned container object automatically is a complex process where huge changes to many libraries are needed and not just an implementation from the consumer side. E.g., Python binding packages such as Pybind11
~\cite{pybind11}
which is aligned more to C++ because it does not expose inner capabilities that Ctype or Cython provides. Doing such management using Ctypes, or Cython requires C-like code in Python.

Moreover, the current protocol demands that all consumer modules create their tensor library and make it visible to Python. However, they should never free the underlying memory pointer when its tensor goes out of scope. But, the tensor library must keep the original destruction function pointer of DLPack Container inside it, and call this function pointer when the tensor goes out of scope. There are a few more steps that are not very clear, hence our attempt to implement this failed.

Lastly, the protocol only mentions that the resultant DLPack tensor should be ``consumed'' once. However, there could be more than one use of any tensor. E.g., the backward computation of SpMM is SpMM and SDDMM, where both operate on the upstream gradient requiring careful understanding and usage of DLPack protocol from the usage side implying whether the upstream gradient be converted to DLPack container twice and used separately.

\subsubsection{\textbf{Proposal: Producer-Only DLPack Protocol}} \label{sec.protocol}
We propose a simpler variant of the DLPack protocol, called producer-only protocol. This is based on our observation about the lifetime of the DLPack container.

\noindent \textbf{Lifetime of DLPack container in a consumer.}
We observe that the job of a consumer such as {\myname-System} is to read from input tensors and write to output tensors, hence the DLPack containers need to remain alive during the System API execution only. This precisely defines the lifetime of ownership by the consumer. However, we observe in Listing~\ref{list-spmm-orig}, the lifetime of \texttt{inDlpackTensor} and \texttt{outDlpackTensor} DLPack containers are also the same as they remain alive till the system API is being executed (Line 13).
\textit{Hence, there seems no need for {\myname-System} to own the DLPack container at all} (Line 10-11). 

\noindent \textbf{Details.}
To this end, we propose a producer-only DLPack protocol where
the producer-owned DLPack container is shared temporarily by {\myname-System} without acquiring its ownership.
{Framework-Adapter} uses the producer APIs to convert the DL Framework tensors to capsule containing DLPack container(Listing~\ref{list-spmm}, Lines 7--8) and then passes the capsule directly to {\myname-System} (Line 10). 
At the end of this Python function, the producer cleans up the DLPack container, as it is about to notice that the container has not been consumed as the value of the container is still ``dltensor''. 
So, capsules remain available for use by the system API.


\begin{lstlisting}[language=Python, caption= {\small Final Attempt: The Framework-Adapter (system integration code). Tensors are allocated in Python, converted to DLPack tensors (capsule), and passed directly to the system API.}, label={list-spmm}]
import GB_System as gbs
import torch as th
def gb_SpMMv(g, inTorchTensor, reduceFn, Forward):
  #Allocate output tensor here and not in Kernel.
  outTorchtensor = th.zeros(g.get_vcount(), featLen)
  #Use Producer API
  inDlpackTensor= th.to_dlPack(inTorchTensor)
  outDlpackTensor= th.to_dlpack(outTorchTensor)
  #calls  SpMMv System API
  gbs.SpMMv(g, indlpackTensor, outDlpackTensor, reduceFn, Forward)  
  return outTorchTensor 
\end{lstlisting}

\vspace{-6pt}

Another point to note here is that the \textit{memory allocation for the output tensor is done in Python} (Line 5, Listing~\ref{list-spmm}) using the DL framework, and never inside the {\myname-System}, and hence we never need to provide the producer side of the protocol at all, thereby resulting into a simple approach of working with DLPack.


\begin{lstlisting}[language=c++, caption= {\small The generated integration code (C++ part) in {\myname} -System for Python binding and DLPack container handling.}, label={list-pybind}]
PYBIND11_MODULE(kernel, m) {
 m.def("SpMMv", [](graph_t& g, py::capsule& in_tensor, py::capsule& out_tensor, int reduceFn, int Forward){
  array_t in_array = capsule_to_array(in_tensor);
  array_t out_array= capsule_to_array(out_tensor);
  return invoke_SpMMv_impl(g, in_array, out_array, reduceFn, Forward);
 });}
\end{lstlisting}

\vspace{-6pt}

Each system API in the {\myname-System} has a Python binding, where it receives the DLPack container capsule, and is part of the critical system integration that needs to extract information from the capsule (Listing \ref{list-pybind}).
We have written a simple $array\_t$ class to keep shape, size, tensor memory pointer, etc. which is filled by reading the capsule, and is passed to the landing code where the final CUDA kernel is invoked. Once we get all the details from the capsule(s), it is not used afterward. 
Hence, when the CUDA kernel stores the results in this memory location during its core computation, the new values are automatically updated in the PyTorch tensor due to the shared reference.


As CUDA kernels are being executed \textit{asynchronously to the CPU}, one may think that the system API (SpMMv) has not been executed fully when the producer has cleaned up the DLPack container. However, this is not a problem, as reading the DLPack container(s) is done in the host (CPU) code that is executed synchronously. 
This leaves our final question regarding the life of underlying memory as one may think the input tensor might be freed once the DLPack container is freed. Even if that happens, it is still not a problem 
as one can also observe the same thing when one follows the consumer side of the protocol, as the consumer will clean up the container even before the CUDA kernel has completed execution. The reason is that the CUDA kernel invocation and GPU memory de-allocation are sent to a queue and will operate in sequence (CUDA semantics). 

\textit{In summary},
The mechanism to pass tensors is almost equivalent to passing tensors as a C-like flat array structure while, of course, using Python capsule.
Since tensor memory is allocated using PyTorch Python API, the whole memory management is still controlled by it, which can de-allocate the memory automatically later without involving {\myname-System}. Even the memory associated with the DLPack tensor container is handled by PyTorch. Therefore, {\myname-System} neither enables python binding for $array\_t$ nor any special handling inside it. The class is a simple C/C++ class.

\subsubsection{\textbf{Significance of Producer-Only DLPack Protocol}}
The significance is listed next.

\noindent \textbf{Powerful System APIs.} 
The protocol makes {\myname-System} independent of the underlying DL framework. Hence, System APIs are as powerful as the interfaces between Python and C++, hence they can accept any custom class as its argument as listed in Table~\ref{table-kernel}. 

\noindent \textbf{Reusability of System Module.}
{\myname} can integrate almost any DL framework that offers conversion of its specific tensor to DLPack tensor. Since PyTorch and Tensorflow offer this, we have validated their integration with {\myname}.
Both the Listings and {\myname-System} are reusable. We require a minor change in the import statement of Listing~\ref{list-spmm} to import TensorFlow and not PyTorch through changes in Line number 6 of the Python code so that tensors can be allocated as per the DL framework. 

Reusability is important for productivity. For example, Meta first implemented submanifold sparse convolution~\cite{graham20183d} using the PyTorch plugin followed by Google~\cite{najibi2020dops} which used the TensorFlow plugin. Google could not reuse the system code-base of Meta because of a different DL framework.
This raises the need for huge investment to make the same functionality available in different DL frameworks.

\noindent \textbf{Applicable to All Consumer System including an Education Tool.}
The proposed protocol has wide-ranging applications in all consumer systems, such as TVM, DGL, etc. The main challenge in all these systems is to access tensors from their system code. Hence, the proposed producer-only protocol applies to many such past and future systems. We also have used the protocol along with  DSL(\cref{sec.dsl}) as an education tool used by a few universities in graduate-level classes to design systems assignments for deep learning. We leave such discussion as future works. 

\subsection{Overview of System APIs} \label{sec.api}
SpMM and SDDMM are two sets of Systems API. however, we choose their variants as system APIs (Table~\ref{table-kernel}), which will be explained soon. Kindly note that GNN also introduces different types of dot product and reduction operators, so we use a generic operator $\odot$ and operator $\oplus$ for their notation which is passed as `eFn' parameter in the APIs. Further, the multi-head attention mechanism in GNN divides each vertex feature of size $|F|$ into a $|H|$ group of $|F'|$ features, where $|H|$ is known as head-count. 


\noindent \textbf{SpMM Variants} always outputs vertex-level tensor.
\textit{SpMMve} is a weighted SpMM where the graph edges contain features(weights). In its forward computation, each destination vertex generates an output ($|F'\times H|$) from features of its source vertices (in-neighbors) ($|F'\times H|$) and edges ($|H|$) using operators $\otimes$ followed by reduction along neighborhood dimension using operator $\oplus$. 
In the variant \textit{SpMMe}, the destination vertex generates output ($|H|$ from features of in-neighbor edges only, each with a feature size of $|H|$. 
The variant \textit{SpMMv} is unweighted, where the destination vertex generates output ($|F'\times H|$) from features of in-neighbor vertices only, each with a feature size of $|F'\times H|$ where $|H|$  is usually 1.
Kindly note that the last two variants do not require the $\otimes$ operator.


\noindent \textbf{SDDMM Variants} output edge-level tensor. In \textit{SDDMMvv}, the features of source and destination vertices, each with $|F'\times H|$ features are operated using operator $\otimes$, and then the intermediate edge results of features $|F'\times H|$ are reduced using operator $\oplus$ in the feature dimension to produce an edge-level tensor, each edge with $|H|$ features. It also has another variant where no operator $\oplus$ is needed.
In SDDMMve, features of source vertices and edges are operated using operator $\otimes$, while it does not need any reduction operator $\oplus$.

\noindent \textbf{Hierarchy of APIs.}
System APIs are meant for system researchers.
{\myname} wraps these System APIs in Python functions, e.g. \texttt{gb\_SpMMv} in Listing~\ref{list-spmm}, and is the lowest level of \textit{Python API} that data scientists use in the operator API's forward() and backward() method (Listing~\ref{list-autograd}). 
Operator APIs and other Pytorch native operators are directly used inside the model layer definition, such as the \textit{GCNConv} layer used in the GCN model. 
The GNN models use such layers. 

\noindent \textbf{Advantage of Fine-grained system APIs.}
Prior GNN systems have designed System APIs on a more need basis resulting in duplication, e.g., GNNAdvisor has designed two sets of SpMM, one for GCN and another for GIN, despite little differences between them. Indeed, many research prototypes~\cite{wang2023tc,huang2021understanding,dgNN,chen2020fusegnn}
have followed such approaches, where APIs have been independently designed for different GNN models. 

Our intuitive system API definition offers various {\myname-system} an opportunity to implement the kernel accordingly, rather than just one SpMM kernel. For example, Cusparse, FeatGraph, Huang et al, etc. offer SpMMve and not SpMMv. The implication is that all these three systems (DGL uses Cusparse) keep a dummy edge-level tensor (O($|E|$) memory) to implement SpMMv using SpMMve leading to additional memory consumption.  Defining SpMMv as the system API will remind future systems to write the correct kernels rather than introducing additional memory overhead, and possibly incurring performance overhead.


There are more advantages of fine-grained system APIs. Consider the case of SpMMv for GCN and one GIN variant. GCN calls normalization by degree array, and the GIN variant uses `mean' as operator $\oplus$. Both of them are the same. Hence, division by degree array happens after SpMMv in the forward computation, hence the same division by degree should happen before SpMMv in the forward computation. Unfortunately, many prior systems compute normalization by degree after SpMM in the backward computation as their System API is not designed to indicate any specific order. Our fine-grained API changes the `norm' flag value passed to System API to indicate if the operation should be done before or after.




\noindent \textbf{Discussion of Message-Passing APIs.}
DGL implements the message-passing APIs in Python on top of operator APIs and is used by GNN model layers. 
Thus message-passing APIs are neither replacements of System APIs nor operator APIs. 
However, its introduction does contribute to framework-runtime overhead.
This overhead has impacted training runtime benchmarking in many prior works that have relied majorly or exclusively on smaller datasets~\cite{wang2021gnnadvisor,fu2022tlpgnn,chen2020fusegnn,gespmmsc2020,dgl2019,wu2021seastar,tian2020pcgcn,dgNN,hu2021efficient,wang2023tc,kim2022analyzing, yan2020characterizing,baruah2021gnnmark,ye2023sparsetir}. 
By directly coming up with an exhaustive list of all the variants, the framework spends no time finding the right System API variant, and hence it avoids that overhead completely. Moreover, the count of variants is only a few, hence users will not have any issues with usability.
Hence, from the benchmarking perspective, we decided not to implement the message-passing.





\subsection{System Integration and Extensibility using DSL} \label{sec.dsl}
Listings~\ref{list-spmm} and \ref{list-pybind}, discussed earlier, represent the system integration code of {\myname} with a DL framework for each system API. Integration-related codes are generated automatically from the API signature to speed up the prototype using our simple domain-specific language (DSL). 
The DSL generates the operator APIs (autograd class), Python wrapper of System APIs, Python binding for System APIs, and a CUDA stub for their implementation.

Such code generation makes {\myname} extensible to support various other forms of systems innovations. One such use case is to support operator or kernel fusion. In such a case, a new System API is needed which is very different from the one listed in Table~\ref{table-kernel}. 
In such a case, the DSL helps with generating all the mentioned code, thus {\myname} speeds up the prototyping.


\noindent \textbf{Mechanism.}
DSL uses API signature written in C++ except that array$\_$t be used to denote the tensor. 
Listing ~\ref{dsl} shows an example of the DSL specification for one kernel.
For the autograd, DSL does not know details about backward computation logic, hence the researcher needs to edit the generated autograd skeleton to call the right kernel(s) in the backward. The system researcher is still responsible for actual CUDA kernel implementation. 

\begin{lstlisting}[language=Python, caption= {\small DSL example for SpMMv system API. We generate the boilerplate code of Listing \ref{list-spmm} and \ref{list-pybind}, and the stub calling CUDA kernel. We also generate autograd skeleton (shown below).}, label={dsl}]
#System API specification in DSL file
status SpMMv(graph_t& g, array_t& input, array_t& output, int reduceFn, int Forward); 
#Generated stub for calling actual C++/CUDA kernel 
status invoke_SpMMv_impl(graph_t& g, array_t& input, array_t& output, int reduceFn, int Forward); 
\end{lstlisting}
\vspace{-6pt}

\begin{lstlisting}[language=c++, caption= {\small The generated Autograd class.}, label={list-autograd}]
class SpMMv(th.autograd.Function):
  @staticmethod
  def forward(ctx, G, X, norm):
    res = gb_SpMMv(G, X, reverse=0, norm)
    ctx.backward_cache=G, norm
    return res 

  @staticmethod
  def backward(ctx, dZ):
    G, norm = ctx.backward_cache
    res = gb_SpMMv(G, dZ, reverse=1, norm)
    return None, res, None
\end{lstlisting}

\vspace{-12pt}

\begin{figure*}[t!]
  \subfloat[Accuracy of original GNNAdvisor and TC-GNN to DGL  \label{exp-accuracy}]{\includegraphics[scale= 0.42]{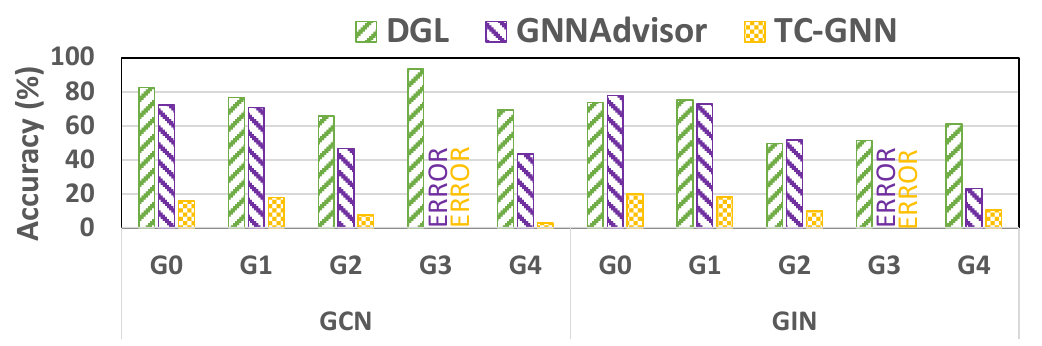}} \qquad
  \subfloat[The accuracy comparison using {\myname} to  DGL. \label{exp-accuracy-new}]{\includegraphics[scale= 0.52]{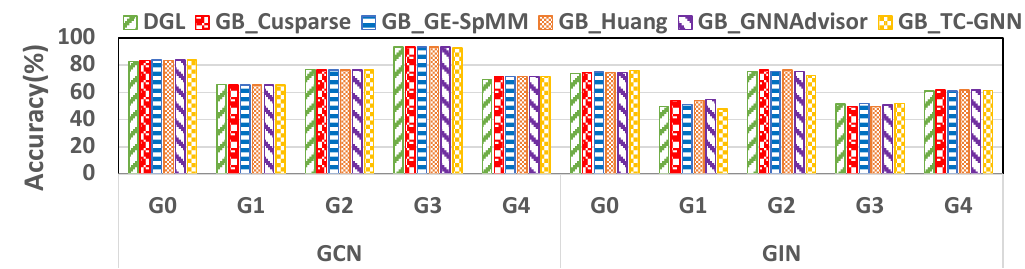}} 
  \caption{\small Results show that {\myname} helps achieve the same accuracy as DGL so can be used for fair evaluations (higher is better)}
\vspace{-12pt}
\end{figure*}


\section{Experiment} \label{sec.exp}

We listed dataset information used for the experiments in Table~\ref{table-dataset}.
Cora, Pubmed, and Citeseer are small datasets. The remaining datasets are referred to as mid-size datasets. 
GCN is a 2-layer model, GIN is a 5-layer model, and GAT is a 3-layer model with intermediate feature-length of 32, 64, and 16 respectively thereby it covers many feature lengths. Other hyper-parameters are kept the same and have been borrowed from DGL.
We run our experiments on one Nvidia A100 GPU with 40GB of device memory.

\subsection{Integrated Benchmarks and Their Significance} \label{sec.significance}
Out of many GNN Systems that have accuracy issues, some of them 
cannot run due to code integration issues, unknown memory corruption, or missing backward computation. So, we picked four top-tier system conferences: GE-SpMM~\cite{gespmmsc2020}, GNNAdvisor~\cite{wang2021gnnadvisor}, Huang et al~\cite{huang2021understanding}, and TC-GNN~\cite{wang2023tc} to not only demonstrate the integration but to generate correctly working artifact for future works by the community. Specifically, GE-SpMM does not work due to integration with DGL due to code reorganization in DGL, and Huang et al do not have backward computation code. GNNAdvisor and TC-GNN show memory corruption when running on some datasets.
They are respectively denoted as GB\_GE-SpMM, GB\_GNNAdvisor, GB\_Huang, and GB\_TC-GNN.


We also built GB\_Cusparse. For GCN and GIN, we use the SpMM kernel of CuSparse. For GAT, we use its SpMM and SpMM$^T$, however, the SDDMMvv of Cusparse is very slow. Hence, except for its SpMM and SpMM$^T$, all other kernels are written by us based on DGL kernel designs.

GB\_Cusparse serves many purposes. \textit{Firstly}, DGL's integration of CuSparse is not optimal DGL manually transposes the edge-level tensor and then calls SpMM to implement SpMM$^T$ despite Cusparse provided a native SpMM$^T$. Hence, advancement by prior works has been shown over a compromised baseline. \textit{Secondly}, DGL also suffers from framework-related memory overhead-- memory that cannot be attributed to any operations, therefore, memory-saving techniques have shown much better results than they should. Hence the community has no baseline information about memory consumption by a GNN system built over Cusparse. 
Worse, DGL has not allowed comparison by prior works in GPUs with less memory capacity due to its frequent out-of-memory.
\textit{Thirdly}, GB\_Cusparse is designed to keep the native performance of Cusparse so that actual runtime and memory improvement over the Cusparse-based GNN system can be measured. We also GB\_Cuspasre to show how much overhead DGL introduces.

Kindly note that many GNN systems only offer SpMM implementations, but not SDDMM, hence they only support some GNN models. 
However, we have automatically extended all GCN-only systems to GIN as they both require the same SpMMv kernel.  Systems such as Huang et al do not have any implementation of SDDMMvv and SpMM$^T$, as they only provide forward computation, hence GAT is not offered.

\noindent \textbf{Outline of Experiments.}
The experiments (\cref{exp-assist-acc}) discusses the accuracy measurements.
\cref{exp-assist-per} shows the impact of framework overhead when prior works have focused on smaller datasets for runtime evaluation, followed by their true comparison using {\myname}. In doing so, we present many new insights and pose new research questions. 
Finally, use dgNN~\cite{dgNN}, a kernel-fusion based system, to compare against GB\_Cusparse to show how much kernel fusion techniques can achieve runtime performance and memory saving so the community can understand the true trade-off in implementation complexity of kernel fusion versus the benefits it brings.

\begin{table}[t]
\footnotesize
\caption{\small Graph datasets. * denotes labeled dataset. GNN models deploy a linear layer to project the feature-length to a lower intermediate feature-length (e.g., 16/32/64) before sparse kernels are called.} 
\centering 
\begin{tabular}{l l r r r r} 
    \hline\hline 
    Graph & Vertex & Edge & Feature & Prediction\\ [0.5ex] 
    Dataset & Count & Count & Length & Class\\ 
    \hline 
    Cora(G0)* & 2,708 &  10,858 & 1,433 & 7\\ 
    Citeseer(G1)*  & 3,327 & 9,104 & 3,703 & 6\\
    Pubmed(G2)* & 19,717 &  88,648 & 500 & 3\\
    Reddit(G3)* & 232,965 & 114,848,857 & 602 & 41 \\
    OGB-Product(G4)* & 2,449,029	& 123,718,280 & 100 & 47 \\
    Amazon(G5) & 400,727 & 6,400,880 & 150 & 6\\ 
    As-Skitter(G6) & 1,696,415 & 22,190,596 & 150 & 6\\
    Cit-Patent(G7) & 3,774,768 & 33,037,894 & 150 & 6 \\ 
    Stackoverflow(G8) & 2,601,977 &	95,806,532 & 150 & 6\\ 
    Hollywood(G9) & 1,069,127 & 112,613,308 & 150 & 6\\
    LiveJournal(G10) & 4,847,571 & 137,987,546 & 150 & 6 \\
    Orkut(G11) & 3,072,627 & 234,370,166 & 150 & 6\\
    
    \hline 
\end{tabular}
\label{table-dataset} 
\end{table}

\begin{figure*}[t!]
  \subfloat[Original Systems \label{exp-small-orig}]{\includegraphics[scale= 0.45]{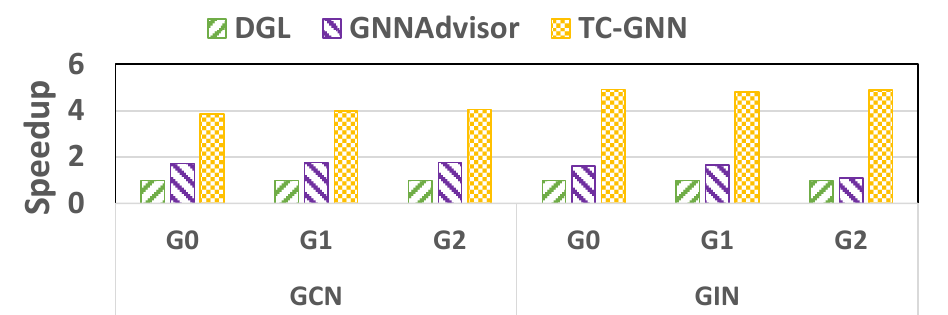}}
  \subfloat[DGL and GB\_* Systems for GCN and GIN \label{exp-small-gcn}]{\includegraphics[scale= 0.43]{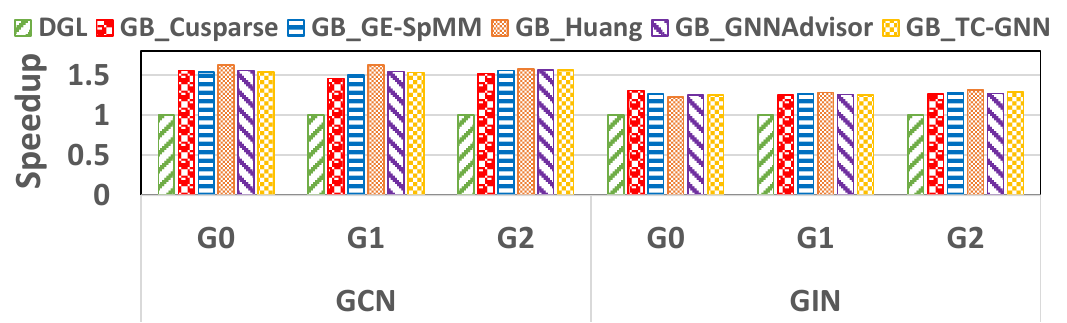}} 
 \subfloat[GAT\label{exp-small-gat}]{\includegraphics[scale= 0.39]{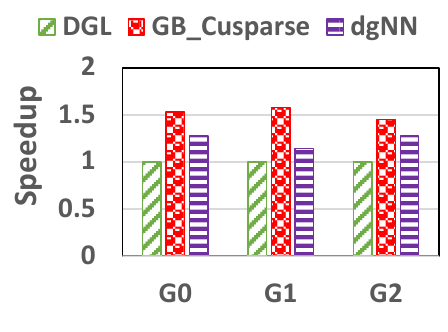}}
  \vspace{-6pt}
 \caption{\small Runtime evaluation on small datasets shows that {\myname} has lower framework overhead than DGL and dgNN (higher is better)}
 \vspace{-18pt}
\end{figure*}


\subsection{Assistance in Identifying Accuracy Issues and Analysis} \label{exp-assist-acc}
Fig.~\ref{exp-accuracy} shows the accuracy of original systems to show their accuracy issues. Fig.~\ref{exp-accuracy-new} presents the accuracy achieved using {\myname}. The analysis is presented next.

\textbf{TC-GNN:} The Original TC-GNN system also has very low accuracy compared to DGL (Fig.~\ref{exp-accuracy}). The loss in accuracy is on average 56.81\%. However, GB$\_$TC-GNN, achieved normal accuracy, as illustrated in Fig.~\ref{exp-accuracy-new}. Further analysis revealed that the original TC-GNN system lacked the normalization function, had no bias operator, and had no support for SpMM with an odd dimension-- needed by SpMM of the last layer, contributing to its abnormal accuracy.


\noindent \textbf{ GNNAdvisor:} Fig.~\ref{exp-accuracy} illustrates that the original GNNAdvisor system achieves abnormal accuracy compared to DGL. {The loss in accuracy is on average 11.89\%, which is very significant, indicating critical issues with the GNNAdvisor system}.
However, GB\_GNNAdvisor achieved normal accuracy (Fig.~\ref{exp-accuracy-new}) implying that integrating with {\myname} automatically solved the accuracy issues. Subsequently, we conducted a detailed code study of the original system, revealing that the abnormal accuracy stemmed from two reasons: a) incorrect backward pass of \textit{normalization by degree array} operator-- the normalization in the backward propagation should precede the SpMM kernel; b) omission of the bias operator.


Our measurements show that the missing normalization and absence of bias parameter degrade the accuracy by {5.1\%, 4.4\%, 0.7\%, 13.1\%, and 20.9\%} G0 to G4 respectively for GCN. \textit{By using {\myname} benchmark platform, users are better equipped to identify and address accuracy issues.}  

GB\_Cusparse, GB\_Ge-SpMM, and GB\_Huang have the same accuracy as DGL (Fig.~\ref{exp-accuracy-new}). 

\subsection{Assistance in Identifying Performance Issues and Analysis}\label{exp-assist-per}
 We show that DGL, the commonly used baseline, incurs significant framework overhead(\cref{exp-dgl-frame}). Consequently, comparing training times with DGL has led to overstated performance speedup as prior works have relied on small datasets. Further, many systems produce errors when running on mid-size datasets, 
so we use {GB\_*} systems for accurate performance evaluations (\cref{exp-assist-true}).


\subsubsection{\textbf{Smaller Datasets and Framework Overhead}} \label{exp-dgl-frame}
GNNAdvisor, TC-GNN, and Ge-Spmm have relied exclusively on smaller datasets for training performance evaluations, while the first two systems have shown better performance than DGL, the latter did not show any. Fig.~\ref{exp-small-orig} confirms that the original GNNADvisor and TC-GNN trains faster than DGL. 
However, benchmarking them using {\myname} (Fig.~\ref{exp-small-gcn}) shows these three systems are not faster than GB\_Cusparse. 


Indeed, GB$\_$Cusparse has 1.51$\times$, 1.32$\times$, and 1.52$\times$ speedup over DGL for GCN, GIN, and GAT models respectively.
The underlying kernels are the same for DGL and GB\_Cusparse, so the results highlight the significant framework overhead inherent in DGL. So, it is clear that the original GNNAdvisor and TC-GNN have lower framework overhead over DGL because these systems utilize custom GNN frameworks that move a lot of Python code to C++ while removing some kernel invocation due to pitfalls. 

Lastly, Ge-SpMM has been integrated into DGL thereby its original evaluation had the same overhead as DGL. 
However, they wrongly blame the last layer of SpMM where the feature dimension is not a multiple of 32.
We could not evaluate the original Ge-SpMM because the integration file has been removed by DGL, making it impossible for us to run. 

Further, these results also show that {\myname} has lower overhead than DGL, and dgNN(\ref{exp-small-gat}). This is because of fine-grained System APIs that we have designed that let us quickly invoke the kernels directly without any other overhead. 


\textit{In summary, evaluation results by the prior works cannot be relied upon when they exclusively used smaller datasets for performance evaluations, where framework overhead dominates the training run time. Integrating into {\myname} provides a fair comparison.}

\begin{figure*}[t!]
  \subfloat[Runtime speedup comparison: DGL is baseline (log-scale, higher is better) \label{exp-large}]{\includegraphics[scale= 0.44]{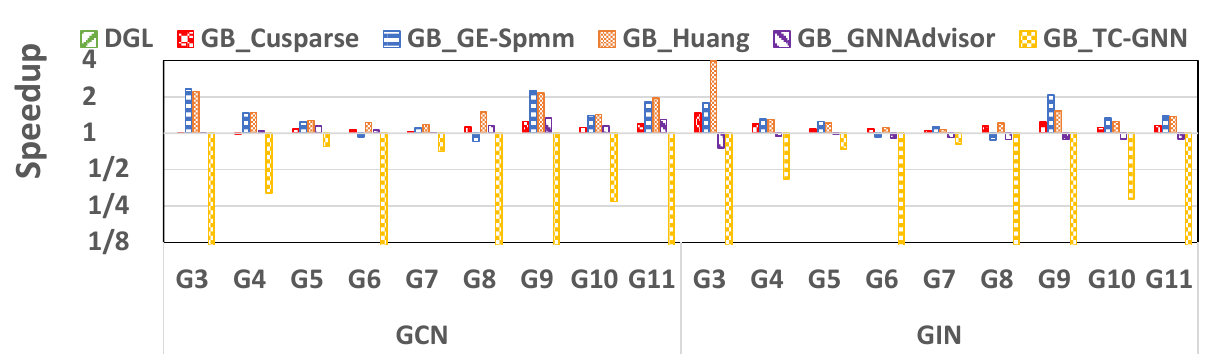}}
   \subfloat[Memory Consumption (lower is better) \label{exp-mem}]{\includegraphics[scale= 0.49]{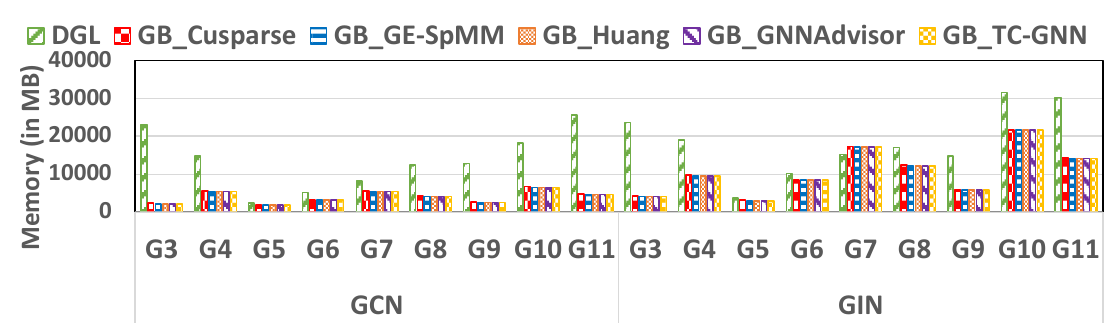}} 
  \vspace{-12pt}
  \caption{\small Runtime performance and memory comparison among {\myname} systems and DGL for GCN and GIN}
\vspace{-12pt}
\end{figure*}




\subsubsection{\textbf{True Comparison of Prior Works}}\label{exp-assist-true}
We evaluate on mid-size datasets to minimize the impact of the framework overhead to obtain accurate runtime comparisons. This way, we can identify which system performs better. However, we use {\myname} for evaluation as Ge-SpMM cannot be compiled due to DGL code reorganization, Huang et al does not have backward computation, while others have memory corruption errors (Fig.~\ref{exp-accuracy}). 
 Fig.~\ref{exp-large} presents our results and shows that TC-GNN is the slowest. GB\_TC-GNN and GB\_GNNAdvisor is 4.24$\times$ and 1.39$\times$ over GB\_Cusparse. Both of them are slower than DGL.
 GE-SpMM and GB\_huang are the the fastest system, achieving on average 1.25$\times$ and 1.39$\times$ speedup over GB\_Cusparse.
 We analyzed the results next.


\textbf{GB\_TC-GNN} 
relies on the TFloat32 data type, which requires 32 bits for storage, the same as the floating-point data type, while internally using the tensor core for computation, which is good when the matrix is dense. However, its tiling structure cannot handle the unstructured sparsity of the graph datasets, and introduces a huge slowdown.   
Kindly note that its original system is the most recently published, but evaluated only on smaller datasets to show better performance.
\textit{To this end, is the tensor core usable for such sparse computations? Asking such questions is possible only due to {\myname} benchmarking and was not answered in the TC-GNN paper.}

\textbf{GB\_GNNAdvisor} 
Kindly note that GNNAdvisor has a workload-balanced solution and borrows caching techniques (of non-zero elements) from Ge-SpMM which does not have any workload balancing. Yet, GNNAdvisor is slower than GE-SpMM apart from being slower than GB\_Cusparse.
The slowdown is due to inappropriate usage of shared memory for the reduction of dot products incurring a total $|E|$ shared-memory writes for reducing dot products, plus an initialization write (of zero) to the shared memory. Indeed, the same functionality can be achieved using a local variable that can reside in registers, as GNNAdvisor never does any inter-thread communication. 
For the GIN model, one SpMMv is performed on the original feature (Table~\ref{table-dataset}), whose length is much larger than the intermediate feature-length. Hence, GNNAdvisor's SpMMv not only incurs large writes to shared memory but also affects occupancy, thus, lowering even more performance. 

\textit{Relying exclusively on smaller datasets misled authors and hence they never noticed the problem to debug the slowdown. }

\textbf{GB\_Huang} also have similar workload balancing technique as GNNAdvisor, yet it performs better than the latter due to register usage for reduction. However, except Reddit, it performs similar to GB\_GE-SpMM while observing a few slowdown. Kindly note that GB\_SpMM does not have any workload balancing.
Huang et al could not exploit the true potential of its workload imbalance because it does not provide a native SpMMv, and rather uses a SpMMve with a dummy edge-tensor whose elements contains 1.0 to simulate it. Due to this additional read of edge-tensor, the performance is not great over GB\_Ge-SpMM.

\textbf{GB\_Cusparse} also achieves an average of 1.13$\times$ for GCN and GIN models over DGL due to overheads of the latter, and other minor engineering practices. E.g., DGL performs degree norm each time by first by first reading  the degree, changing the 0 degree to 1, performing 1 over degree, and then multiplying with features. {\myname} caches the final inversion of degree array.  

\textit{In summary, the results question prior works and show that {\myname} can help in fair evaluation. 
}

\subsection{Framework-Memory Overhead and Consumption Analysis}\label{true-memory}
Fig.~\ref{exp-mem} shows that DGL consumes huge memory over GB\_Cusparse despite both systems use SpMM of Cusparse. Fig.~\ref{exp-gat-large} shows similar results for the GAT model.
While DGL consumes additional memory for its graph storage and has other memory-intensive designs as it keep edge ID for GAT kernels; these do not account for such huge memory differences. For example, when measured at kernel level for Reddit, DGL's SpMMv (feature-length = 32) consumes 7,276 MB of memory alone, while GB\_Cusparse consumes around 936 MB proving that the input vertex-tensor, the dummy edge-tensor, and the output vertex-tensor cannot consume this much memory in DGL. \textit{Thus, DGL introduces a memory overhead of around 6,340 GB of memory.} Kindly note that DGL's SDDMM does not introduce memory overhead consuming exactly 936 MB of memory for input and output tensors.

 Our analysis shows that DGL uses producer-consumer DLPack protocol, but internally it brings PyTorch dependency for its internal memory allocation, which is needed for Cusparse SpMM's scratch pad buffer. Howeer, DGL's SDDMM does not require any such buffer. Thus, we suspect the internal integration to be a reason for the overhead. The issue has been conveyed to the DGL team.


Fig.~\ref{exp-mem} also show that GB\_* systems consume similar memory 
with only minor memory consumption differences, as some of the system have storage metadata, and dummy edge-tensor that consumes very little additional memory.

  
\begin{figure}[t]
\vspace{-6pt}
\includegraphics[scale= 0.50]{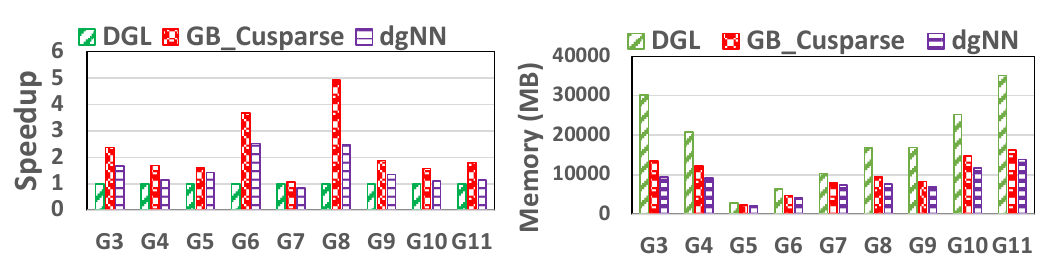}
\vspace{-18pt}
\caption{\small Runtime speedup (higher is better)  and memory consumption (lower is better) comparison among DGL, GB\_Cusparse, and dgNN for GAT on mid-size datasets.} 
\label{exp-gat-large}
\end{figure}

\subsection{GAT Analysis and Kernel Fusion Advantage(?)}
Fig.~\ref{exp-gat-large} shows that GB\_Cusparse achieves an average 2.38$\times$ speedup and 39.40\% memory saving over DGL.
 This is because of the advantage of using native SpMM and SpMM$^T$ by GB\_Cusparse in the forward and backward pass over DGL which uses eShuffle+SpMM in both the forward and the backward computation. The eShuffle shuffles the edge-level tensor using edge ID, which was introduced because DGL may be unaware that native SpMM$^T$ is present in Cusparse.  

We choose dgNN, a GNN system specialized in kernel fusion for attention-based GNNs, such as GAT, and does not support GCN and GIN.
However, dgNN is slower by 45.77\% and saves only 16.44\% memory over GB\_Cusparse.
The result shows that the kernel fusion done by dgNN cannot bring any performance advantage over a baseline that utilizes the Cusparse kernels natively. Our analysis indicates that currently studied kernel fusion techniques use vertex-centric parallelism while fusing SpMM and SDDMM. This leads to workload imbalance in the fused kernels. On the other hand, GB\_Cusparse uses COO for SDDMM, as it mimics DGL, achieving workload balancing. 

Kernel fusion increases the effort on the implementation as custom kernels are written on a per-model basis rather than having generic kernels, such as SpMM and SDDMM. 
\textit{These results make the community more informed about whether to spend effort on kernel fusion or not and force us to re-visit and re-evaluate the kernel fusion in GNN.}











\subsection{Versatility of {\myname}}

{\myname} can integrate with multiple DL frameworks, such as Pytorch, Tensorflow, and MXNet easily. While the majority of our experimental results are evaluated using the PyTorch integration, we also demonstrate compatibility with TensorFlow for GCN to show that integration with any general-purpose DL framework is possible that offers DLPacktensors. Table.\ref{table-graphpy-TFvsPT} shows that Pytorch and Tensorflow integration achieve similar accuracy.

\begin{table}[h]
\small
\vspace{-6pt}
\caption{\small {\myname}-TensorFlow vs {\myname}-PyTorch accuracy comparison for GCN by integrating Cusparse.} 
\vspace{-6pt}
\centering 
\begin{tabular}{|l| r| r|} 
    \hline\hline 
    Graph Dataset & {\myname}-Tensorflow & {\myname}-Pytorch\\ [0.5ex] 
    \hline 
    Cora & 83.18\% &  83.40\%\\ 
    Citeseer & 65.32\% & 65.10\%  \\
    PubMed & 76.47\% &  76.52\%\\
    Reddit & 92.89\% &  92.96\%\\
    OGB-Product & 69.76\%& 69.67\%\\
    \hline 
\end{tabular}
\vspace{-6pt}
\label{table-graphpy-TFvsPT} 
\end{table}

\section{Conclusion} \label{sec.conclusion}
In conclusion, {\myname} presented a benchmarking platform for productive research and fair evaluations, which unearthed intriguing findings and new artifacts.
Further, the producer-only DLPack protocol and generation of critical system integration code allow systems researchers to productively prototype any novel system ideas amid low framework overhead and no framework-memory overhead while enabling many other use cases, which we will explore in the future. 

\balance


\bibliographystyle{ieeetr}
\bibliography{refer/GraphPy}


 


\end{document}